\documentclass[11pt,a4paper]{article}
\usepackage[hyperref]{acl2019}
\usepackage{times}
\usepackage{latexsym}
\usepackage{url}
\usepackage{pst-node}
\usepackage{mathtools}
\usepackage{amssymb, amsmath}
\usepackage{tikz}
\usetikzlibrary{topaths}
\usepackage{array, multirow, boldline}
\usepackage{mathtools}

\usepackage{float}
\restylefloat{table}

\usepackage[caption=false]{subfig}
\usepackage{hyperref}

\aclfinalcopy 
\def\aclpaperid{104} 

\title{Transfer Learning from an Auxiliary Discriminative Task for Unsupervised Anomaly Detection}

\author{Urwa Muaz 
\\
  Center for Urban Science and Progress \\ New York University\\
  \texttt{urwa.muaz@nyu.edu} \\\And
  Stanislav Sobolevsky\\
  Center for Urban Science and Progress\\
  New York University,\\
  Institute Of Design And Urbanism\\
ITMO University, Saint-Petersburg\\
  \texttt{sobolevsky@nyu.edu}}

\date{}
\tikzstyle{every picture}+=[remember picture,inner xsep=0,inner ysep=0.25ex]

\begin{document}
\maketitle
\begin{abstract}
Unsupervised anomaly detection from high dimensional data like mobility networks is a challenging task. Study of different approaches of feature engineering from such high dimensional data have been a focus of research in this field. This study aims to investigate the transferability of features learned by network classification to unsupervised anomaly detection. We propose use of an auxiliary classification task to extract features from unlabelled data by supervised learning, which can be used for unsupervised anomaly detection. We validate this approach by designing experiments to detect anomalies in mobility network data from New York and Taipei, and compare the results to traditional unsupervised feature learning approaches of PCA and autoencoders. We find that our feature learning approach yields best anomaly detection performance for both datasets, outperforming other studied approaches. This establishes the utility of this approach to feature engineering, which can be applied to other problems of similar nature.  

\end{abstract}

\section{Introduction}
Recent availability of big data on human mobility broadens our horizons of understanding of human society at global \cite{hawelka2014geo,belyi2017global} and local scale \cite{kung2014exploring, kang2013exploring, amini2014impact} and urban transportation in particular \cite{santi2014quantifying,nyhan2016predicting,tachet2016revisiting}. However complexity and dimensionality of the data along with sparsity of available measurements at the local scale and resulting high noise-to-signal ratio challenges our ability of detecting robust and meaningful patterns from it.

Unsupervised anomaly detection is a major frontier of machine learning research with widespread applications in many domains, transportation being one of them. There is an increasing interest in detection of anomalous mobility patterns and congestion events. According to the Federal Highway Administration (FHWA) these non recurring congestion events account for approximately 55 \% of delays caused in travel times of the drivers in the United States \citet{1}. Predicting future traffic congestions can inform the route planning and scheduling and prevent worsening of these conditions. Anomalies in urban mobility patterns can also be indicative of potentially dangerous situations. Examples of these scenarios include 2015 New Year’s Eve celebrations in Shanghai, where overcrowding resulted in a stampede causing 36 casualties. Early detection of these events can enable the authorities to take preventive measures to mitigate and prevent the consequences.

The core of anomaly detection is  constructing a probabilistic model of normal behavior and identifying anomalies as observations with small likelihood under the model. Mobility network data is usually very high-dimensional and traditional anomaly detection methods do not perform well with high dimensional data, especially when the dataset size is not many times larger than the number of dimensions \citet{2}. This is a generic issue with machine learning models and is referred to as the curse of dimensionality \citet{3}. Most prior works address this issue by adopting a two stage approach \citep{4,2}, where low dimensional representation is learned prior to applying anomaly detection techniques on the latent representation. 

Unsupervised feature learning from unlabelled data is itself an important area of research. Traditionally, low dimensional representation is achieved using statistical decomposition methods like PCA \citet{5} or deep representation learning methods like auto-encoders \citet{6}. PCA can be used to encode the data into low dimensional space with aim to capture most of the variance of the raw data, and auto-encoders compress the data into low dimensional representation space with the objective of accurate reconstruction of the raw data. In this study we formulate an arbitrary classification task from the data for which labels are known and train deep neural networks classifiers on it. We then investigate if the deep features extracted from the higher layers of this classification network are transferable to the unsupervised anomaly detection task. Furthermore, we evaluate this approach on anomaly detection in urban mobility and compare it to other feature learning approaches mentioned above. We find that transfer learning from an auxiliary classification task to unsupervised task on the same data set is a promising approach and yields better results than other feature learning approaches considered in this study.

Our main contributions are summarized as follows: i) we show that discriminative training on an auxiliary classification task is a promising approach to learn meaningful feature representation which is transferable to unsupervised anomaly detection, this has not been studied before to the best of our knowledge ii) we provide an empirical comparison of our approach with existing approaches of PCA and autoencoders iii) we validate our approach on a downstream application of anomaly detection in urban mobility networks iv) we show that in presence of underlying topology, use of graph convolutional layers give better feature space as measured by efficacy of anomaly detection.

\begin{table*}[h]
    \centering
    \begin{tabular}{c c c c c c c c c}
    \hline\hline\\[-2.5ex]
    City & Days & Nodes & Avg. Daily Ridership & National Holidays \\ 
    \hline
    Taipei & 638 & 108 & 112055 & 30 \\
    New York & 548 & 263 & 7060 & 15 \\
    \end{tabular}
    \caption{Dataset Summary}
    \label{tab:table1}
\end{table*}

\begin{table*}[h]
    \centering
    \begin{tabular}{c c c c c c c c c}
    \hline\hline\\[-2.5ex]
     &  \multicolumn{2}{c}{New York} & \vline & \multicolumn{2}{c}{Taipei} \\ 
    
    Network & Nodes & Edges & \vline & Nodes & Edges \\
     \hline
    Raw & 263 & 65722 & \vline & 108 & 11664\\
    Aggregated & 24 & 576 & \vline & 10 & 100 \\
    \end{tabular}
    \caption{Topological Aggregation of Mobility Network}
    \label{tab:table2}
\end{table*}

\section{Related Work}

The most widely used method for anomaly detection is density estimation of the feature space and isolating low probability density observations as anomalies. Methods used for density estimation include multivariate Gaussian Models multivariate Gaussian Models, Gaussian Mixture Models, k-means \citep{7,8,9}. Since the network data is usually high-dimensional, direct application of these algorithms suffer from the curse of dimensionality. Most prior works address this issue by adopting a two stage approach \citep{2,4}, where low dimensional representation is learned from high dimensional data first, and subsequently density estimation based anomaly detection is applied to this latent representation. There are numerous methods in literature to achieve low dimensional representation from data.

Statistical decomposition methods perform dimensionality reduction by leveraging the fact that usually high dimensional data has underlying low dimensional structure and thus most of the variance of the data can be encoded in a few dimensions. Most widely used means of achieving matrix decomposition is Principal Component Analysis (PCA) \citet{5}, and its more sophisticated variants have also been studied \citep{10,11}. We use PCA as one of our baseline methods.

Unsupervised Deep representation learning methods like autoencoders \citet{6} and generative adversarial networks (GAN) \citet{12} are used to learn a latent representation that are capable of accurate reconstruction or generation of data. Unlike linear PCA they are capable of learning complex non-linear relationships. Autoencoders have been widely used in computer vision to learn powerful representations from unlabelled data \citep{2,17,13,14,15}. They are also a popular dimensionality reduction technique for anomaly detection in high dimensional dataset \citep{16,17}. \citet{18} uses a coupled pipeline of deep autoencoder and Gaussian Mixture Models for unsupervised anomaly detection. \citet{35} builds upon this to proposes a three stage pipeline approach for anomaly detection in high dimensional network data sets, they use topological aggregation before Autoencoder and Gaussian Mixture Models to tackle the issue of high noise to signal ratio in edge level measurements. Some previous works use cost associated with reconstruction as feature for anomaly detection. They rely on the assumption that anomalies can not be accurately reconstructed from the representation space, which does not always hold so latent space provides more reliable features \citet{18}. Similarly to autoencoders, use of GAN as a representation learning technique is also a growing research field \citep{19,20}. We implement deep autoencoders as a second baseline for comparison with our proposed methodology.

Discriminative training can be used for feature learning from unlabelled data by creating an auxiliary classification task to train neural networks. Activations from deeper layers can be used as a low dimensional latent space for end application. Their use in computer vision have shown that they are capable of learning powerful representations. Examples of this type of tasks in computer vision are learning the relative positions of image patches \citep{21,22} , colorizing grayscale images \citep{23,24}, or learning the geometric transformations applied on images \citet{8}. Usually, discriminative feature learning is used as a pre-training method for transfer learning into another supervised task. First research that investigates the use of features learned from a supervised task for unsupervised learning problem is conducted by \citet{25}. This research shows that performance of deep CNN features from imagenet for image clustering is comparable to other state of the art unsupervised feature learning methods. To the best of our knowledge the transferability of features learned from classification to unsupervised anomaly detection have not been studied before. In this research we investigate this problem.

\section{Methodology}

\subsection{Datasets}
\textbf{Mobility Datasets}: In this study we used two publicly available real world urban mobility datasets. First dataset is a subway ridership dataset for the city of Taipei, this data contains origin destination mobility for 108 subway stations. 21 months of data ranging from January 2017 to September 2018 was used in this study. Second dataset is taxi trips for the city of New York, this contains origin destination trip information for 263 taxi zones in New York. 18 Months of data ranging from July 2017 to December 2018 was used for this study. Both datasets were aggregated at a daily level for experiments. Since ridership counts over time could be affected by data collection which sometimes changes over extensive periods of time, we apply normalization along the spatial axis to avoid temporal inconsistencies.

\textbf{Events Datasets}: We are investigating a novel methodology for unsupervised anomaly detection but we need a mechanism to compare the performance of our method to other existing techniques. For that purpose, we collected a dataset of dates which we believe will have anomalous urban mobility patterns. National holidays are a good choice because they result in closing of public and private institutions and disrupt the commute patterns of the city. Furthermore, they occur more frequently as compared to other rare events, rendering the performance estimates more reliable. There were 30 national holidays in Taipei and 16 in New York for the respective durations under study. Dataset summary is provided in table \ref{tab:table1}.

\subsection{Pipeline Approach}
A two staged pipieline approach is common in prior works \citep{2,4}, where first stage is feature learning where a meaningful low dimensional representation is learned from high dimensional data, and second stage is application of density estimation based anomaly detection to this feature representation. \citet{35} proposes a three staged pipeline for anomaly detection in high dimensional network data, they add a topological aggregation as a first step before feature learning to tackle the issue of high noise to signal ratio in edge level measurements. The approach has been further evaluated in \citet{36} for anomaly detection in urban mobility across several major cities, and yielded superior results to other approaches under study. We adopt this three stage pipeline approach when dealing with edge level features. Within this anomaly detection pipeline framework we test our proposed feature learning method as its second step and compare it with existing methods by evaluating the overall performance of the resulting pipeline.

\subsection{Network Representation}
The mobility data with origin destination information is a natural candidate for being represented as a temporal network. Each day of data is represented as a graph, where Taxi zones in New York and subway stations in Taipei become the nodes of the network. Two different network configurations have been used for this research, first models ridership as node features and the other regards them as edge features. 

\textbf{Edge Features}: In this configuration there are no node features and edge features denote the ridership between the nodes. The edge level features have very high dimensionality and also suffer from high noise to signal ratio, so we need to use some sort of aggregation.  We use the pipeline approach proposed by \citet{35}. Topological aggregation of network is performed through community detection to address these issues prior to low dimensional feature engineering. Community aggregation also reduces the dimensionality of the graph data, making it more manageable for experiments. We use COMBO proposed by \citet{26} for community aggregation of the network in this study. The effect of network aggregation on network size is summarized in table \ref{tab:table2}. This pipeline approach is used for all experiments and baselines for edge feature configuration.

\textbf{Node Features}: Alternatively we aggregate the incoming and outgoing traffic for each node to get the node features. This reduces dimensionality of the problem allowing to focus on a key characteristic of mobility such as local distribution of its volume. It might be efficient for many applications discovering patterns where the local volume is impacted and individual nodes have sufficient data volume, however missing the network structure could be seen as a limitation. However the advantage of this representation is that one can use it along with the actual physical connections of the transport system. This allows us to use further graph convolution localizing the filters based on transport connections. Such physical connections are clearly represented between subway stations unlike the taxi zones, so this network representation is only used for Taipei subway data. 

\begin{figure}[ht]
\centering
\includegraphics[width=70mm]{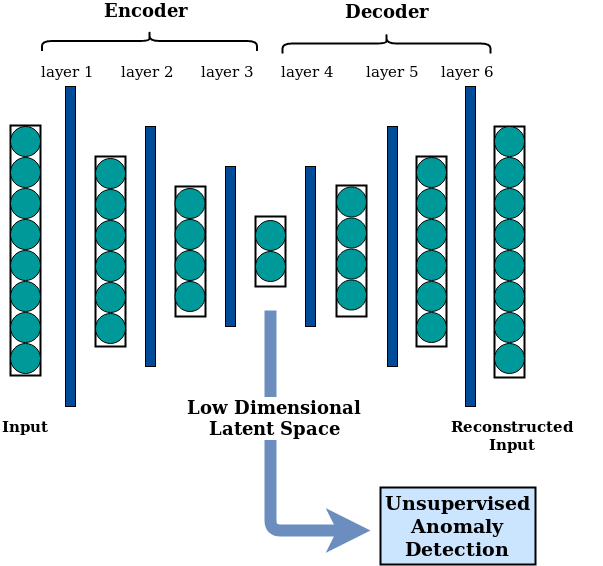}
\caption{Transfer Learning from AutoEncoder}
\label{fig:ae}
\end{figure}

\begin{figure}[ht]
\centering
\includegraphics[width=62mm]{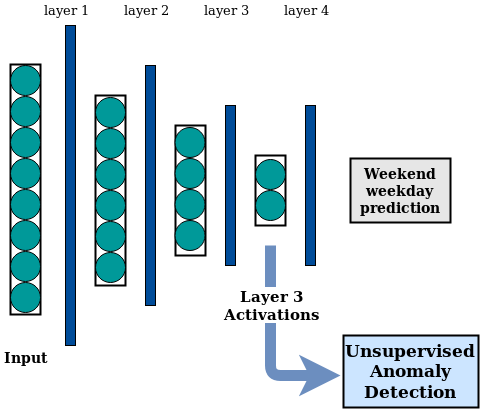}
\caption{Transfer Learning from Auxiliary Classification}
\label{fig:ann}
\end{figure}

\subsection{Feature Learning}

Since the network data is high-dimensional, direct application of anomaly detection algorithms suffer from the curse of dimensionality \citet{2}. Prior works address this issue by adopting a two stage approach \citep{2,4}, where low dimensional representation is learned from high dimensional data first, and subsequently density estimation based anomaly detection is applied to this latent representation. Thus, to further reduce the complexity of high-dimensional network data enabling the application of anomaly detection, we need to learn a suitable low dimensional features space. A large body of traditional dimensionality reduction tools are available, most common ones being statistical decomposition methods like PCA \citet{5} and deep representation learning methods like auto-encoders \citet{6}. Along with our proposed feature learning approach we also implement traditionally used techniques from the literature. All the methods represent the data into 20 dimensional latent space for fair comparison. 

We propose an auxiliary classification task which we believe will enable the classifier to learn representations that will be useful for anomaly detection. We train models to discriminate weekdays from weekends based on daily ridership patterns. We use a Multilayered perceptron (MLP) composed of four fully connected layers, where the number of nodes in each layer decrease steadily to achieve dimensionality reduction while preserving the information necessary for classification. Figure \ref{fig:ae} and Figure \ref{fig:ann} provide a architectural comparison of our approach and deep autoencoders, one of the baselines. Network is trained for the classification task and the activations of third layer are used as a 20 dimensional latent space. This does not use topological structure on the network. This method is denoted as \textbf{Discriminative-MLP} in the results. Model trained on cross entropy loss using adam optimizer. 100 epochs of training are performed with weight decay and learning rate of 0.001.

\textbf{Graph Convolutions:} Node feature representation of Taipei also encodes the topological information of connectivity of stations. Transport networks are known to have strong spatial and structural correlations. Multilayered percepton is very good at modeling complex non linear relationships but it does not adequately address the spatial and structural dependencies between different nodes in a traffic network accurately. Convolutional neural networks (CNN) based approaches \citep{27,28}  have been used in transport networks because of their ability to model spatial relationships between nodes. Though, traditional CNN work well with modelling images and other spatial relationships in Euclidean space, they are not appropriate for networks where the connectivity and structure goes beyond spatial proximity. Recently, convolution operators have been generalized for graph domain and many variants exist in the literature. Recently, they have been applied to traffic modelling problems with great success \citep{33,34}.
We use first order approximations of spectral graph convolution introduced by \citet{29} in this study to model spatial relationships. Higher order convolutions can be achieved by stacking multiple first order graph convolution layers. We experiment with preceding the MLP with graph convolutional layers to see if this improves the quality of latent representations. In this model the MLP component operates on latent representation of nodes rather than actual node features. This architecture is referred to as \textbf{Discriminative-GCN} in the results. We stacked different number of first order GCN layers to find the optimal order of convolution for anomaly detection in Taipei subway network. Training parameters and method is similar to MLP method described above.

\begin{table*}[h!]
    \centering
    \begin{tabular}{c c c c c c c c c}
    \hline\hline\\[-2.5ex]
    Features & F1 Score & Precision & Recall & Holidays Identified \\ 
    \hline
    PCA & 0.458 & 0.579 & 0.379 & 11 \\
    Autoencoder & 0.433 & 0.419 & 0.448 & 13 \\
    Discriminative MLP & 0.575 & 0.451 & 0.793 & 23 \\
    Discriminative GCN & \textbf{0.633} & 0.655 & 0.613 & 19 \\
    
    \end{tabular}
    \caption{Anomaly detection results for Node Feature representation of Taipei}
    \label{tab:table3}
\end{table*}

\begin{table*}[h!]
    \centering
    \begin{tabular}{c c c c c c c c c}
    \hline\hline\\[-2.5ex]
    Features & F1 Score & Precision & Recall & Holidays Identified \\ 
    \hline
    PCA & 0.547 & 0.455 & 0.690 & 20 \\
    Autoencoder & 0.550 & 0.431 & 0.759 & 22 \\
    Discriminative MLP & \textbf{0.778} & 0.840 & 0.724 & 21 \\
    
    \end{tabular}
    \caption{Anomaly detection results for Edge Feature representation of Taipei}
    \label{tab:table4}
\end{table*}

\begin{table*}[h!]
    \centering
    \begin{tabular}{c c c c c c c c c}
    \hline\hline\\[-2.5ex]
    Features & F1 Score & Precision & Recall & Holidays Identified \\ 
    \hline
    PCA & \textbf{0.563} & 0.563 & 0.563 & 9 \\
    Autoencoder & 0.200 & 0.130 & 0.438 & 7 \\
    Discriminative MLP & \textbf{0.563} & 0.563 & 0.563 & 9 \\
    
    \end{tabular}
    \caption{Anomaly detection results for  New York}
    \label{tab:table5}
\end{table*}

\subsection{Feature Learning Baselines}
\textbf{PCA} essentially learns a linear transformation that projects the data into another space, where vectors of projections are defined by the variance of the data. By restricting the dimensionality to a certain number of components that account for most of the variance of the data set, we can achieve dimensionality reduction. We retain first 20 components to achieve a 20 dimensional latent representation.

\textbf{Autoencoders} are neural networks that can be used to reduce the data into a low dimensional latent space. As shown in figure \ref{fig:ae} they have a encoder-decoder architecture, where the encoder maps the input to latent space and decoder reconstructs the input. They are trained using back propagation for accurate reconstruction of the input. By intuition, these low dimensional latent variables should encode most important features of the input since they are capable of accurate reconstructing of the input. We implement autoencoder to represent our data into 20 dimensional latent space. Both the encoder and decoder are composed of three layers and the network is trained using binary cross entropy loss. 100 epochs of training are performed with weight decay and learning rate of 0.01 using adams optimizer.

\subsection{Density Estimation}
Gaussian Mixture Model \citet{31} is used to construct a density estimation model of normal behavior and anomalies are identified as observations with small probability density under this model. A simple approach would be to fit a multivariate Gaussian to data and perform outlier detection based on p-value threshold as done by \citet{30}. But real world data often has multiple underlying distributions and assumption of a single distribution does not hold. In our case, the weekdays are expected to have a different ridership distribution than weekends. A potential solution is to use gaussian mixture models (GMM) \citet{31}, which is parametric probability distribution model which represents the data distribution as weighted sum of normally distributed sub-populations. Thus we use GMM for density estimation and outlier detection is then performed by p-value thresholding on component sub-populations as done by \citet{32}.

\subsection{Model Evaluation}
It is hard to form a robust evaluation mechanism for unsupervised anomaly detection since the ground truth is not known. We use the ability of the methods to identify National Holidays from normal days as a proxy of its performance. It is worth noting that we should not except very high performance scores in these experiments since National Holidays do not constitute all the anomalies and the false positives we get might actually be other events resulting in anomalies in transport networks. For convenient comparison of the approaches described above, we report the best F1 score for each of the methods. F1 score gives a one value performance metric by computing harmonic average of precision and recall.

\section{Results and Discussion}
The results for the experiments are summarised in table \ref{tab:table3} and table \ref{tab:table4}. As it can be observed that representations learned from discriminative training on auxiliary classification task provides best F1 scores across all experiments. For node feature representation of Taipei data discriminative features perform significantly better than both PCA and autoencoders, and using graph convolution layers further improve the anomaly detection F1 scores. Similarly, for edge feature representation experiments, discriminative feature learning outperforms traditional techniques. Thus, it can be empirically backed that discriminative training using an auxiliary classification task is a viable approach for feature learning and these features are transferable to unsupervised tasks like anomaly detection. it is interesting to note that for New York the performance of anomaly detection is significantly lower than Taipei. National holidays might have a stronger effect on the mobility patterns of Taipei than New York, or the changes might be reflected more strongly in subway ridership than taxi ridership. Furthermore, since Taipei has twice the number of national holidays so its results are statistically more reliable.

Feature learning methods achieve low dimensional latent representation with different objectives, which might not result in representations suitable for unsupervised anomaly detection. PCA performs dimensionality reduction with aim to retain most of the variance of the data but still the information vital for anomaly detection might reside in the omitted information, resulting in inferior performance in anomaly detection. Autoencoder compresses the data into latent space that is capable for accurate reconstruction of the output, but the features that are essential for anomaly detection might not be essential for reconstruction. We note that the auxiliary classification task that we proposed was closely related to the original task of anomaly detection, hence transferability of features was high. If latent space has enough information to separate weekdays and weekends it is plausible that it is also good at distinguishing holidays from normal days. 

Finally, we note that use of GCN improves the quality of latent representations. GCNs are able to model spatial and structural dependencies by using localized filters, so that is why they were able to learn more stable representations for Taipei subway network, which has a clearly defined topology.  

\section{Conclusion}

This paper evaluated the tranferrability of features learned from an auxiliary classification task to unsupervised anomaly detection in temporal networks, using human mobility as an example.The proposed discriminative feature learning is used within a suitable pipeline approach for anomaly detection as one of its phases reducing overall dimensionality of the task. 
The results clearly indicate the usefulness of this approach to attain efficient feature learning by improving the overall anomaly detection F1 scores over the traditional unsupervised feature learning approaches. 

In broader terms, this work further establish the tranferrability of features from supervised learning to unsupervised tasks such as anomaly detection. We believe that this approach of finding a suitable classification task which has available labels and perform supervised classification to extract features, is highly generalizable to other domains where the labelled data is not available. Our results establish it as promising alternative to unsupervised and self supervised feature learning techniques. 

We have seen that discriminative training results in best representations for our unsupervised learning task but it is critical to note that transferability of the features would depend on the nature of the classification task and its relevance to the downstream unsupervised task. Identification of a suitable classification task with the availability of labels can be seen as a limitation of this approach.

In future works, it would be interesting to investigate how the transferability of the features changes with the nature of the task. 

\bibliographystyle{acl_natbib}
\nocite{*}
\bibliography{acl2019.bib}

\end{document}